\documentclass[pdflatex,sn-mathphys-num]{sn-jnl}% Math and Physical Sciences Numbered Reference Style 
\usepackage{graphicx}%
\usepackage{multirow}%
\usepackage{amsmath,amssymb,amsfonts}%
\usepackage{amsthm}%
\usepackage{mathrsfs}%
\usepackage[title]{appendix}%
\usepackage{xcolor}%
\usepackage{textcomp}%
\usepackage{manyfoot}%
\usepackage{booktabs}%
\usepackage{algorithm}%
\usepackage{algorithmicx}%
\usepackage{algpseudocode}%
\usepackage{listings}%
\usepackage{makecell}

\begin{document}

\title[Article Title]{LLMs in Political Science: Heralding a New Era of Visual Analysis}


\author{\fnm{Yu} \sur{Wang}}\email{yuwang.aiml@gmail.com}


\affil{\orgdiv{Institute for Advanced Study in Social Sciences}, \orgname{Fudan University}, \orgaddress{\city{Shanghai}, \postcode{200433}, \country{China}}}

% \affil*[2]{\orgdiv{School of Journalism}, \orgname{Fudan University}, \orgaddress{\city{Shanghai}, \postcode{200433}, \country{China}}}

% \affil[3]{\orgdiv{Department}, \orgname{Organization}, \orgaddress{\street{Street}, \city{City}, \postcode{610101}, \state{State}, \country{Country}}}


\abstract{Interest is increasing among political scientists in leveraging the extensive information available in images. However, the challenge of interpreting these images lies in the need for specialized knowledge in computer vision and access to specialized hardware. As a result, image analysis has been limited to a relatively small group within the political science community. This landscape could potentially change thanks to the rise of large language models (LLMs). This paper aims to raise awareness of the feasibility of using Gemini for image content analysis. A retrospective analysis was conducted on a corpus of 688 images. Content reports were elicited from Gemini for each image and then manually evaluated by the authors. We find that Gemini is highly accurate in performing object detection, which is arguably the most common and fundamental task in image analysis for political scientists. Equally important, we show that it is easy to implement as the entire command consists of a single prompt in natural language; it is fast to run and should meet the time budget of most researchers; and it is free to use and does not require any specialized hardware. In addition, we illustrate how political scientists can leverage Gemini for other image understanding tasks, including face identification, sentiment analysis, and caption generation. Our findings suggest that Gemini and other similar LLMs have the potential to drastically stimulate and accelerate image research in political science and social sciences more broadly.}




\maketitle

% unsupervised_semi_supervised_visual_frames,face_detection_political_figures_news_archives,video

% finetune_pa,cross_domain
Political scientists are now in an era of data abundance, and machine learning tools are increasingly used to extract meaning from data sets both textual and visual~\citep{mlss}. While the primary emphasis has thus far centered on textual data~\citep{finetune_pa,cross_domain}, more and more political scientists are endeavoring to utilize image data in their analyses~\citep{learn_to_see}. Images play an important role in the political world, whether it is election campaigns, military conflicts, immigration, or poverty. As artificial intelligence continues to advance and its adoption expands, political scientists are progressively utilizing AI to analyze the composition and impact of visual materials in political contexts~\citep{unsupervised_semi_supervised_visual_frames,face_detection_political_figures_news_archives}. Typically, this requires political scientists to have substantial training in computer vision and machine learning more broadly. For example, researchers need to understand concepts in computer vision such as the FAST Hessian detector~\citep{hessian}, SIFT features~\citep{sift}, R-CNN features~\citep{rcnn}. Not to mention the fact that many of these features and models nowadays require the use of specialized hardware such as graphics processing units (GPUs). Stringent requirements of computer vision expertise coupled with hard requirement on hardware have made progress on image analysis in political science particularly slow. Even though it is widely accepted that images contain vast amount of valuable information, so far only a selected few political scientists are capable of carrying out research in this field. Their research outputs, in turn, can only reach a rather small audience due to the technicality of computer vision studies and the difficulty in reproducing and further advancing the results.

This  paper  explores  the  potential  of  large  language  models  (LLMs)  for  image understanding tasks, with a focus on Gemini, which was released in December 2023 by Google. It demonstrates that zero-shot image annotation by Gemini has an average rating of 3.8 on a 4-point Likert scale, with 1 being poor and 4 being excellent. While a few recent works have shown that LLMs can act as a valuable tool in text annotation~\citep{zero-shot,gpt-hate}, this paper shows how LLMs can perform common image understanding tasks, such as object detection, face identification and captioning, to facilitate image studies in political science and social sciences more broadly. We focus on the task of object detection because political scientists have a strong need to understand what objects exist in the images for improved interpretability and downstream modeling and because object detection has long been a bottleneck for political scientists~\citep{unsupervised_semi_supervised_visual_frames,face_detection_political_figures_news_archives}. We use 688 images from 33 news outlets. For each image, we use Gemini to extract the objects contained in the image and grade the identified objects from 1 to 4. Our results show that Gemini does a good job in identifying objects from images. The quality is sufficient for downstream tasks such as topic modeling~\citep{stp} and ideology classification. Moreover, throughout the entire process, the only command we used was a simple prompt asking Gemini to retrieve objects, and we did not use any specialized computer vision knowledge. This should help remove the technical barrier for all political scientists. Moreover, the cost of using Gemini is zero, and it does not require any specialized hardware on the researchers' side. Lastly, in terms of latency, it is relatively fast: it took 3,800 seconds to process 688 images or 5.5 seconds per image. This should meet the time budget of most researchers. Researchers who require higher throughput could, for example, use multiple processes.

\section*{Results}
We utilize the dataset from~\cite{unsupervised_semi_supervised_visual_frames}, which focuses on the Central American migrant caravan and is recently published at the leading political science journal \textit{Political Analysis}. The dataset consists of 688 images collected from 33 news outlets. The distribution of images among these outlets is uneven. For example, there is one 1 from Axios and Buzzfeed each, 45 from USA Today and 46 from Vice News (Table~\ref{result}).

\begin{table}[!b]
\caption{Average annotation ratings, rounded to the closest decimal, for all outlets across the entire political spectrum. The average ratings fall between 3 and 4 for all 33 news outlets.}
\label{result}
\setlength{\tabcolsep}{15pt}
\begin{tabular}{lclc}
\hline\hline
News Outlet               &  \# Images & Political Leaning & Avg. Rating \\\hline
ABC News                  & 26               & Left-Center        & 3.9            \\
Al Jazeera English        & 38               & Center             & 3.7            \\
Associated Press          & 25               & Center             & 3.5            \\
Axios                     & 1                & Center             & 3.0            \\
Bloomberg                 & 3                & Center             & 3.7            \\
Breitbart News            & 31               & Right              & 3.8            \\
Business Insider          & 45               & Center             & 3.7            \\
Buzzfeed                  & 1                & Left-Center        & 4.0            \\
CBS News                  & 20               & Left-Center        & 4.0            \\
CNBC                      & 43               & Center             & 3.9            \\
CNN                       & 25               & Left-Center        & 3.7            \\
Fortune                   & 6                & Not-Rated          & 4.0            \\
Fox News                  & 23               & Right-Center       & 3.8            \\
MSNBC                     & 2                & Left               & 3.5            \\
Mashable                  & 18               & Left               & 3.9            \\
NBC News                  & 11               & Left-Center        & 3.9            \\
National Review           & 41               & Right              & 3.9            \\
New York Magazine         & 20               & Left               & 3.6            \\
Newsweek                  & 34               & Left-Center        & 3.4            \\
Politico                  & 16               & Left-Center        & 3.9            \\
Reuters                   & 6                & Center             & 3.8            \\
The American Conservative & 5                & Right-Center       & 3.0            \\
The Hill                  & 22               & Center             & 3.7            \\
The Huffington Post       & 31               & Left               & 3.7            \\
The New York Times        & 18               & Left-Center        & 3.7            \\
The Verge                 & 4                & Left-Center        & 4.0            \\
The Wall Street Journal   & 34               & Center             & 3.7            \\
The Washington Post       & 12               & Left-Center        & 4.0            \\
The Washington Times      & 18               & Right-Center       & 3.9            \\
Time                      & 15               & Left-Center        & 3.9            \\
USA Today                 & 45               & Center             & 3.7            \\
Vice News                 & 46               & Left               & 3.6            \\
Wired                     & 3                & Not-Rated          & 4.0   \\\hline
\textit{All (weighted)}                     &    688             &           & 3.8 \\\hline\hline
\end{tabular}
\end{table}

We use Gemini to annotate each image and manually evaluate Gemini annotations using a four-point Likert scale extending from 1 (poor), 2 (average), to 3 (good) and 4 (excellent) regarding their overall quality, with the understanding that a grading of 3 would indicate that Gemini annotations are potentially useful for conducting research on political images~\citep{gpt4_orthopedic_treatment}. When grading somewhat fits between two scores, we assign it the lower score to be more conservative. Across the entire corpus, we observe that out of 688 images, 593 (86\%) receive a rating of `excellent', 49 (7\%) receive a rating of `good', 14 (2\%) a rating of average and 32 (5\%) `poor'. The average rating across the entire corpus is 3.8.

% When the grading falls somewhat between two scales, we round it down to be more conservative.
% 

%We evaluate Gemini's performance at three levels: the image level, the news outlet level, and the corpus level.

%Outperforms humans on this: politico\_0006\_3.jpg

In total, the model used 504 words (or phrases), the most frequent word is `person', showing up in 349 out of 688 images, and the least frequent words, such as `shipping container', `surgical mask' and `graduation cap', are only used once. Words that are often associated with the migrant caravan, such as `blanket' (24 times) and `stroller' (13 times), are also included in the vocabulary. Interested readers could refer to the replication materials for the details of the vocabulary and its distribution.

% \raisebox{2\height}{4: Excellent}& \makecell[l]{`person': 1 \\`microphone': 1\\\\\\}& \raisebox{-.2\height}{\includegraphics[width=0.2\textwidth,height=0.12\textwidth]{al-jazeera-english_0000_1.jpg}} \\ 
% \hline 

% \raisebox{2\height}{4: Excellent}& \makecell[l]{`person': 15\\ `truck': 1\\ `blanket': 1\\ `phone': 1\\\\}& \raisebox{-.27\height}{\includegraphics[width=0.2\textwidth,height=0.12\textwidth]{associated-press_0007_7.jpg}} \\ 

\begin{table}[h] 
\centering
\caption{For each given image, we first retrieve the objects using Gemini calls, and then evaluate the detected objects on a four-point Likert scale: poor, average, good, and excellent.}\label{sample}
\setlength{\tabcolsep}{26pt}
\begin{tabular}{|l|l|c|}
\hline
Rating& Objects& Images\\\hline

\raisebox{2\height}{4: Excellent}& \makecell[l]{`person': 1 \\`microphone': 1\\\\\\}& \raisebox{-.2\height}{\includegraphics[width=0.3\textwidth,height=0.16\textwidth]{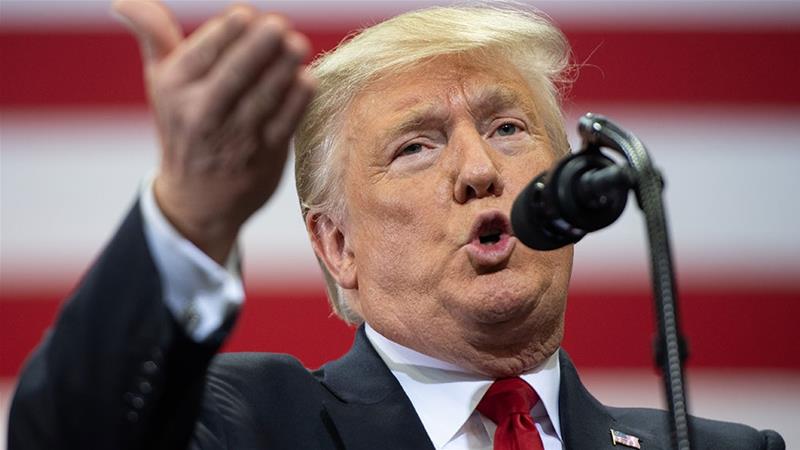}} \\ 

\hline 

\raisebox{1\height}{3: Good}& \makecell[l]{`person': 1\\ `flag': 1\\ `hat': 1\\ `shirt': 1 \\\\}& \raisebox{-.3\height}{\includegraphics[width=0.3\textwidth,height=0.16\textwidth]{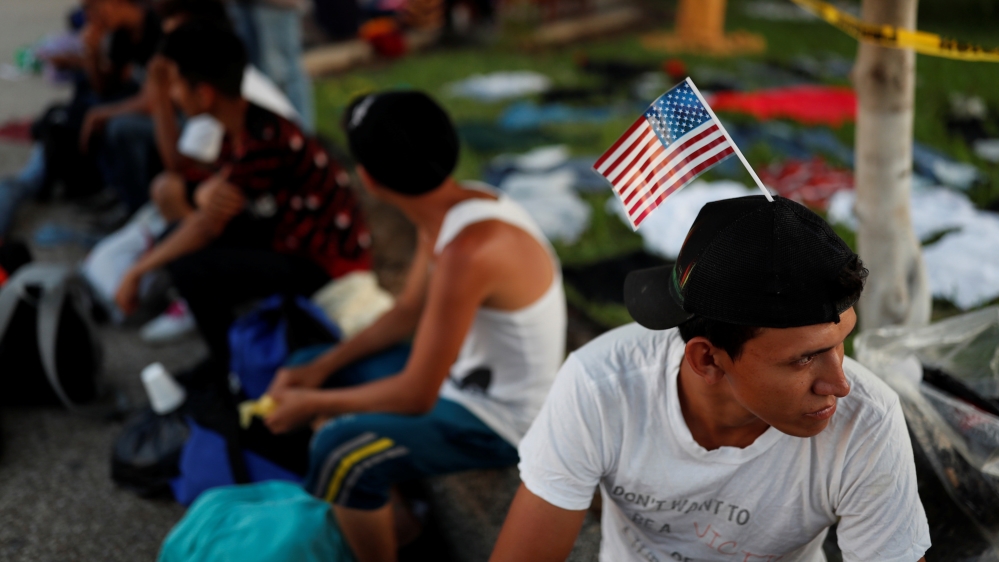}} \\ \hline 

2: Average& \makecell[l]{ `city': 3\\ `state': 3\\ `country': 3 \\`line': 1  \\`map': 1\\ \\} & \raisebox{-.34\height}{\includegraphics[width=0.3\textwidth,height=0.16\textwidth]{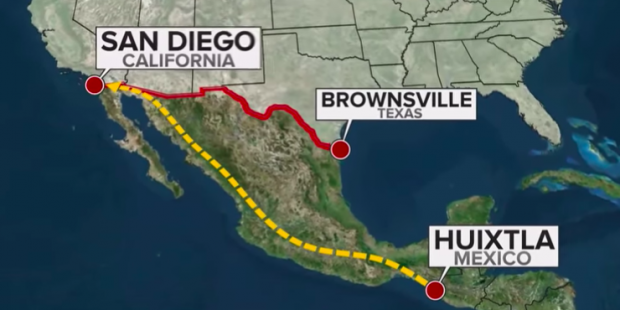}} \\ \hline 

 \raisebox{2\height}{1: Poor}&  \makecell[l]{\textit{None}\\\\\\\\} & \raisebox{-0.2\height}{\includegraphics[width=0.3\textwidth,height=0.16\textwidth]{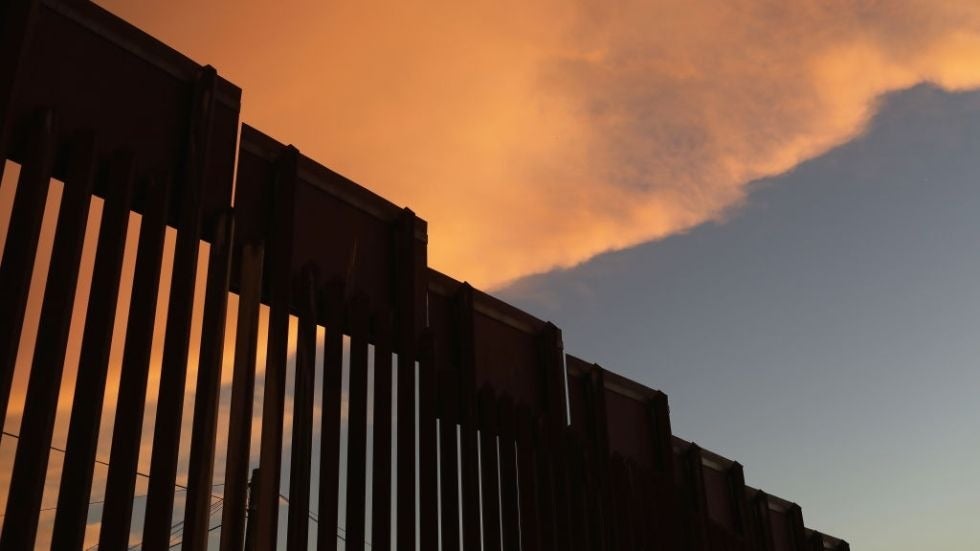}} \\ 
\hline 

\end{tabular} 
\end{table}

We note that Gemini tends to perform particularly well with images with one or two persons. Most of these images are leading U.S. politicians such as Donald Trump, Nancy Pelosi, and Mike Pence. Gemini tends to perform poorly on images that are either screenshots or figures with a large amount of text.  In Table~\ref{sample}, we present some images, their respective ratings as well as detected objects and their counts. To view the detected objects for all the images, interested readers could refer to the replication materials.

%At the corpus level, we observe that out of 688 images, 593 (82\%) receive a ration of `excellent', 49 (7\%) receive a rating of `good', 14 (2\%) a rating of average and 32 (5\%) `poor' (Figure~\ref{fig:1}). The average rating across the entire corpus is 3.75, which should meet the needs of most image researchers in political science.

% * Number of vocabulary is 2,000. 

% * Max number of objects

% * Max number of types of objects.

% * Randomness

% * Add all the gradings

% \begin{figure}[h]
% \centering
% \includegraphics[height=8.5cm]{rating.pdf}
% \caption{This legend would be placed at the side of the figure, rather than below it.}\label{fig:1}
% \end{figure}

%In Table~\ref{result}, we report the evaluation results. We observe that ...

\section*{Discussion}
 Our study shows that Gemini excels in delivering highly accurate object detection for images on the issue of Central American migrants. The results from Gemini were accurate, capturing most objects in an image as well as providing a reliable estimate of the count for each object. We conclude that Gemini's performance is impressive, particularly considering that its annotations are zero-shot. 
 
 Note that we have mainly focused on object detection, because it is arguably the most common and fundamental task for image analysis. Once detected, these objects can then be used for downstream tasks such as clustering, topic modeling, and ideology classification. This does not mean, however, that Gemini's capabilities are limited to object detection. In Figure~\ref{more_image_tasks}, we further illustrate how political scientists can leverage Gemini to retrieve information about known politicians, sentiment undertone, and captioning of a given image.

 %  Given that different research projects, e.g. image captioning, may have different requirements for accuracy and prompting, we make the image dataset~\citep{unsupervised_semi_supervised_visual_frames}, detected objects, detailed annotation, and our code available, so that researchers could conveniently evaluate for themselves whether Gemini, as an alternative to training one's own computer vision models, could be adequate for their image related projects in political science.

\begin{table}[h] 
\centering
\caption{Gemini can be used for various image tasks, including face identification, sentiment undertone detection, and captioning, all of which are of significant value to political scientists.}\label{more_image_tasks}
\setlength{\tabcolsep}{2pt}
\begin{tabular}{|l|p{2.8cm}|p{3cm}|c|}
\hline
Image Tasks& Prompts& Gemini & Images\\\hline

% {\includegraphics[width=0.3\textwidth,height=0.16\textwidth]{cbs-news_0000_9.jpg}} \\ 

{Face identification}& List the known politicians in the image. & The image shows Nancy Pelosi, the Speaker of the United States House of Representatives.\newline & \raisebox{-.7\height}{\includegraphics[width=0.3\textwidth,height=0.16\textwidth]{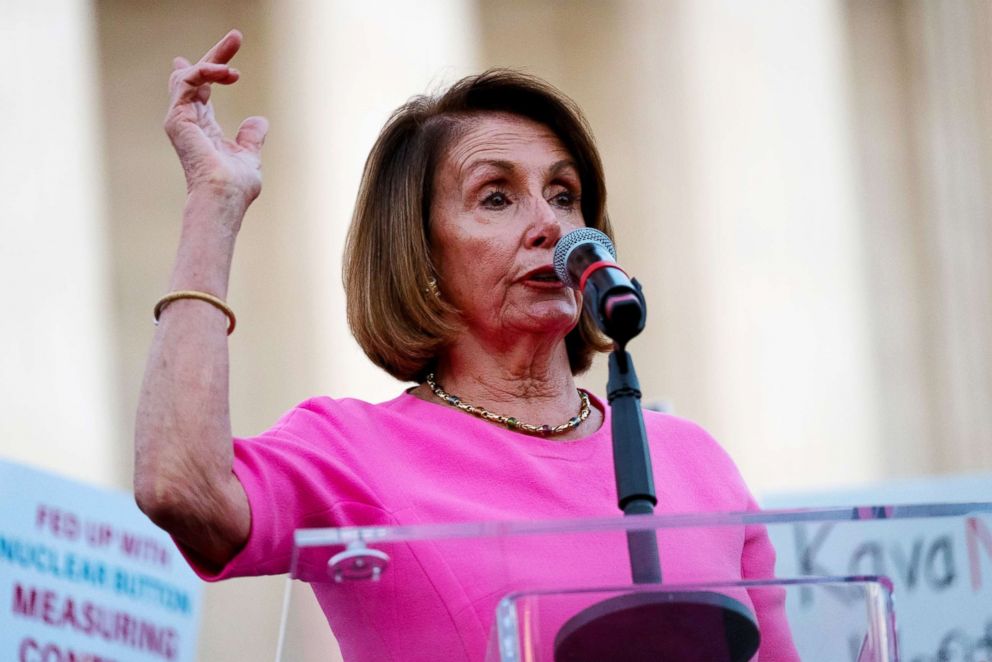}} \\ 

\hline 

 Sentiment undertone &{Use up to four words to describe the sentiment undertone of the image.\newline\newline} & Serious, vigilant, watchful, cautious. &\raisebox{-.7\height}{\includegraphics[width=0.3\textwidth,height=0.16\textwidth]{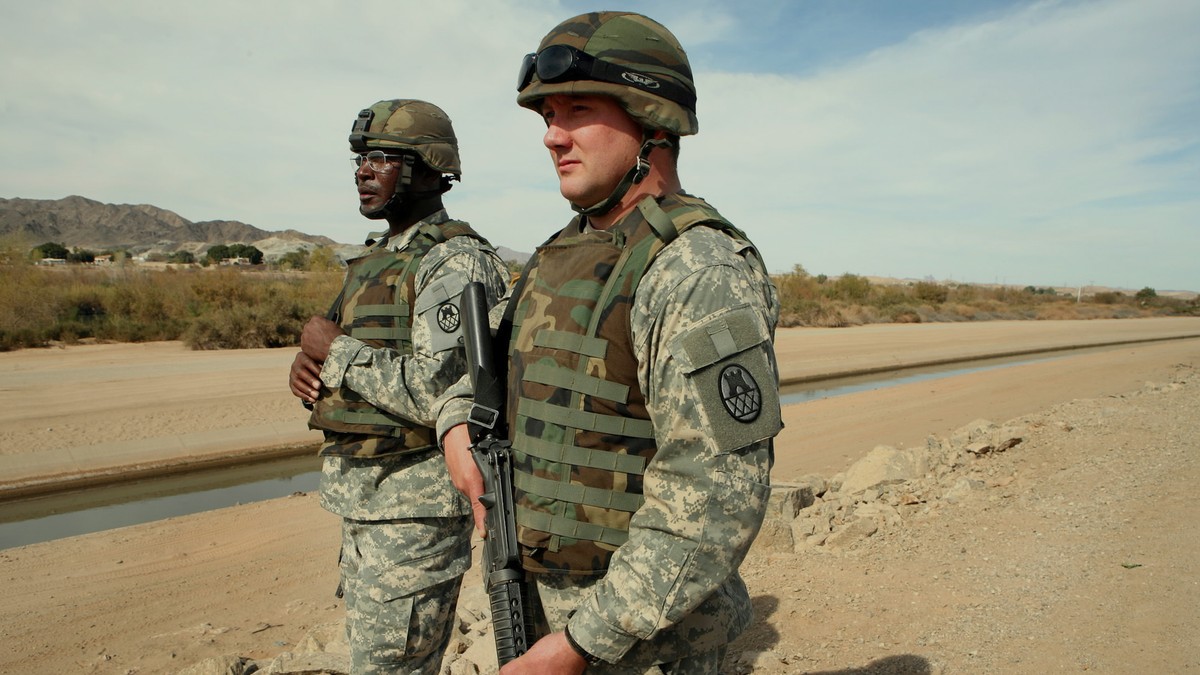}} \\ \hline 

 Image captioning& Generate a brief caption for the given image.  &  A group of protesters hold signs and march down a street demanding the reunification of families that were separated at the border.\newline &\raisebox{-.8\height}{\includegraphics[width=0.3\textwidth,height=0.16\textwidth]{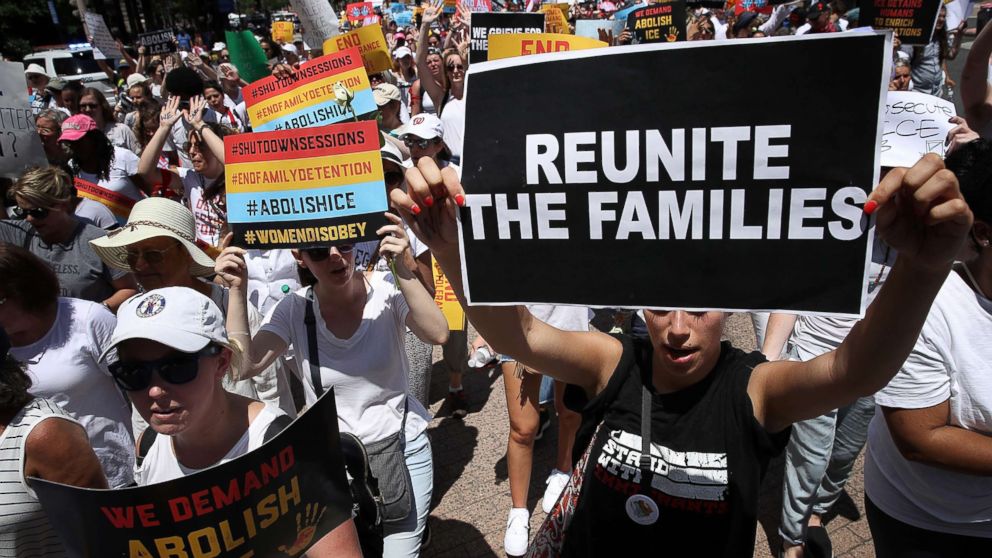}} \\ 
\hline
\end{tabular} 
\end{table}

% ~\citep{prompt}
 Besides its impressive performance, we note three other advantages of leveraging Gemini for image research. One is that Gemini's latency is fairly low at 5.5 seconds per image, which translates to 650 images per hour. Another is that Gemini is free to use. Researchers would not need access to specialized hardware, such as high memory GPUs. A third advantage is that calling Gemini requires minimal coding and machine learning expertise. All it takes is a prompt in natural language that reflects the needs of the researcher. We are confident that with the help of Gemini and other LLMs the majority of political scientists should be able to conduct image analysis whether their substantive interests lie in American politics, comparative politics, or international relations.

\section*{Materials and methods}
\subsection*{Dataset Characteristics}
The analysis relies on a corpus of 688 images in 424 news articles from 33 news outlets covering the caravan of Central American migrants. In terms of political leaning, the news outlets cover the entire spectrum, ranging from Left (e.g., MSNBC) to Left-Center (e.g., CNN), Center (e.g., USA Today), Right-Center (e.g., The Washington Times), and to Right (e.g., Breitbart News). The news articles were published between October 3 and November 1, 2018~\citep{unsupervised_semi_supervised_visual_frames}.

\subsection*{Gemini Encoding and Prompting}

Gemini was accessed online on January 14th, 2024, and operated as the gemini-pro-vision version. Prompts were provided in a standardized format. Note that we instruct Gemini to return the analysis results (1) as a one-line dictionary, (2) in lowercase, and (3) in singular forms to facilitate easier data analysis downstream.

\begin{itemize}
  \item List all the objects and people in the image, using lowercase and singular forms. Include the count for each object. Please use the following one-line format with curly brackets in the response: \{``Object 1'': Count 1, ``Object 2'': Count 2\}.
 
\end{itemize}

Researchers interested in different facets of image analysis, such as captioning and sentiment analysis, are advised to create and test distinct prompts tailored to their specific needs. Our focus on object detection is mostly to showcase Gemini's capability of solving common computer vision problems in political science.

%\noindent List all the objects and people in the image. Include the count for each object. Please use the following one-line format with curly brackets in the response: \{``Object 1'': Count 1, ``Object 2'': Count 2\}

\subsection*{Grading Rubrics}
%We describe the details of the grading rubrics in Table~\ref{rubric}. When Gemini does not find any objects or blocks the processing of an image, we assign it a one. When Gemini captures some information, we assign it a two. When several objects are correctly identified, we assign a three. Lastly, when most information is correctly identified, we assign it a four. Note that we do not require the object counts to be exactly correct for a rating of four.

In this subsection, we describe the details of the grading rubrics. When Gemini does not find any objects or erroneously blocks the processing of an image, we assign it a one (poor). We also assign it a one when it grossly misrepresents the image. When Gemini captures some information from the image, we assign it a two (average). When several objects are correctly identified but there are noticeable gaps, we assign a three (good). Lastly, when most of the relevant information is correctly identified, we assign it a four (excellent). Note that we do not require the object counts to be exactly correct for a rating of four.

% \begin{table}[!h]
% \caption{Grading rubrics for overall quality of image analysis reports.}
% \label{rubric}
% \begin{tabular}{lll}
% \hline\hline
% Scale & Answer    & Description                                       \\\hline
% 1     & Poor      & Grossly misrepresent the image                    \\
% 2     & Average   & Capture some information of the image             \\
% 3     & Good      & Capture a good amount of information in the image \\
% 4     & Excellent & Captures most information in the image  \\\hline
% \end{tabular}
% \end{table}

\subsection*{Data, Materials, and Software Availability}
Replication materials will be made available  at  the  Harvard  Dataverse, including all the images, annotations, ratings, as well as the code.

\bibliography{ir}% common bib file

\end{document}